\def\year{2017}\relax
\documentclass[letterpaper]{article}
\usepackage{aaai17}
\usepackage{times}
\usepackage{helvet}
\usepackage{courier}
\usepackage{graphicx}
\usepackage{amsfonts}
\usepackage{subfig}
\newcommand{\dcr}{DCR}

\newcommand*\samethanks[1][\value{footnote}]{\footnotemark[#1]}

\begin{document}

%Title still needs work
%Mo's version
% \title{End-to-End Reading Comprehension with Dynamic Answer Chunk Ranking}
\title{End-to-End Answer Chunk Extraction and Ranking for Reading Comprehension}
% A Neural Dynamic Answer Chunk Ranker for Reading Comprehension
% Neural Answer Chunking and Ranking for Reading Comprehension
% Dynamic Chunk Reader for Reading Comprehension
% Reading Comprehension for Factoid and Non-Factoid Questions Answering with Dynamic Chunk Ranking
% \title{An End-to-End Neural Networks to Rank Dynamic Chunks for Reading Comprehension}
\author{
Yang Yu\thanks{Both authors contribute equally}, Wei Zhang\samethanks, Kazi Hasan, Mo Yu, Bing Xiang, Bowen Zhou\\
\{yu, zhangwei, kshasan, yum, bingxia, zhou\}@us.ibm.com\\
IBM Watson
}

\maketitle
\begin{abstract}

This paper proposes {\it dynamic chunk reader} ({\it DCR}), an end-to-end neural reading comprehension (RC) model that is able to extract and rank a set of answer candidates from a given document to answer questions. DCR is able to predict answers of variable lengths, whereas previous neural RC models primarily focused on predicting single tokens or entities. DCR encodes a document and an input question with recurrent neural networks, and then applies a word-by-word attention mechanism to acquire question-aware representations for the document, followed by the generation of chunk representations and a ranking module to propose the top-ranked chunk as the answer. Experimental results show that DCR achieves state-of-the-art exact match and F1 scores on the SQuAD dataset \cite{rajpurkar2016squad}.

\end{abstract}

\section{Introduction}

%Kazi
Reading comprehension-based question answering (RCQA) is the task of answering a question with a chunk of text taken from related document(s). A variety of neural models have been proposed recently either for extracting a single entity or a single token as an answer from a given text \cite{hermann2015teaching,kadlec2016text,trischler2016natural,dhingra2016gated,chen2016thorough,sordoni2016iterative,cui2016aoa}; or for selecting the correct answer by ranking a small set of human-provided candidates \cite{yin2016attention,trischler2016parallel}. In both cases, an answer boundary is either easy to determine or already given.

%Owing to the nature of the proposed RC datasets \cite{hermann2015teaching,weston2015towards,hill2015goldilocks,richardson2013mctest,tapaswi2015movieqa,voorhees2000trec,hewlett2016wikireading,onishi-16}, deep learning models capable of answering both factoid and non-factoid questions were not explored until the recent release of the Stanford Question Answering Data (SQuAD) dataset \cite{rajpurkar2016squad}.
Different from the above two assumptions for RCQA, in the real-world QA scenario, people may ask questions about both entities (factoid) and non-entities such as explanations and reasons (non-factoid) (see Table~\ref{example} for examples). 
%TODO By Mo: I list two versions for comparison. I feel like (1) you hope to make the discussion focusing more on non-factoid, which demonstrates the strength of RC. However, first it ignores the strength of our system on factoids; and second we are not that good on non-factoids so it is difficult to make the points (2) I delete the Baidu paper because it focuses on factoids and added the SQuAD citation
In this regard, RCQA has the potential to complement other QA approaches that leverage structured data (e.g., knowledge bases) for both the above question types. This is because RCQA can exploit the textual evidences to ensure increased answer coverage, which is particularly helpful for non-factoid answers.
However, it is also challenging for RCQA to identify answer in arbitrary position in the passage with arbitrary length, especially for non-factoid answers which might be clauses or sentences.
% In this regard, we believe that RCQA can complement other QA approaches that leverage structured data (e.g., knowledge bases), as the non-factoid answers are more likely to be found in text. However, extracting non-factoid answers is a challenging task for RCQA systems as the answer chunks are of variable lengths.
%% we believe that for both the above tasks text-based RCQA can be an important complementary to other QA approaches leveraging structured data (e.g., knowledge bases), as in many cases the answers (especially the non-factoid ones) are more likely to be found in text.
%% We believe that text-based RCQA has advantages over approaches leveraging other forms of structured data on both the above cases, as the non-factoid answers are more likely to be found in text rather than in structured data forms (e.g., knowledge bases). 
%% However, the above goal also raises new challenges for RCQA methods to find from document the answer chunks in arbitrary positions with arbitrary lengths. 
As a result, apart from a few exceptions \cite{rajpurkar2016squad,wang2016machine}, this research direction has not been fully explored yet. 

%By Mo: (3) answer positions and bounaries sounds similar. Also after we mentioned learning ... separately or jointly, this sentence is not connected the the next ones. (4) we did not define cloze style before
Compared to the relatively easier RC task of predicting single tokens/entities\footnote{State-of-the-art RC models have a decent accuracy of $\sim$70\% on the widely used CNN/DailyMail dataset \cite{hermann2015teaching}.}, 
predicting answers of arbitrary lengths and positions significantly increase the search space complexity:
% predicting answers of arbitrary lengths involves learning both answer positions and answer boundaries either separately or jointly. 
the number of possible candidates to consider is in the order of $O(n^{2})$, where $n$ is the number of passage words. In contrast, for previous works in which answers are single tokens/entities or from candidate lists, the complexity is in $O(n)$ or the size of candidate lists $l$ (usually $l\leq$5), respectively.
To address the above complexity, Rajpurkar et al.~\shortcite{rajpurkar2016squad} used a two-step chunk-and-rank approach that employs a rule-based algorithm to extract answer candidates from a passage, followed by a ranking approach with hand-crafted features to select the best answer. The rule-based chunking approach suffered from low coverage ($\approx$ 70\% recall of answer chunks) that cannot be improved during training; and candidate ranking performance depends greatly on the quality of the hand-crafted features. More recently, Wang and Jiang \shortcite{wang2016machine} proposed two end-to-end neural network models, one of which chunks a candidate answer by predicting the answer's two boundary indices and the other classifies each passage word into answer/not-answer. Both models improved significantly over the method proposed by Rajpurkar et al.~\shortcite{rajpurkar2016squad}.

Our proposed model, called {\it dynamic chunk reader} ({\it DCR}), not only significantly differs from both the above systems in the way that answer candidates are generated and ranked, but also shares merits with both works.
First, our model uses deep networks to learn better representations for candidate answer chunks, instead of using fixed feature representations as in \cite{rajpurkar2016squad}.
Second, it represents answer candidates as chunks, as in \cite{rajpurkar2016squad}, instead of word-level representations \cite{wang2016machine}, to make the model aware of the subtle differences among candidates (importantly, overlapping candidates). 
% In this way, our end-to-end neural networks model adopts a candidate ranking approach \cite{rajpurkar2016squad}. 
% While Wang and Jiang~\shortcite{wang2016machine} also uses neural networks to learn word representations, they did not model candidate chunks explicitly. 

%By Mo: passage encoder not defined before
%modeled, which could jeopardize the model effectiveness. 

The contributions of this paper are three-fold. (1) We propose a novel neural network model for joint candidate answer chunking and ranking, where the candidate answer chunks are dynamically constructed and ranked in an end-to-end manner. (2) we propose a new question-attention mechanism to enhance passage word representation, which is subsequently used to construct chunk representations. (3) We also propose several simple but effective features to strengthen the attention mechanism, which fundamentally improves candidate ranking, with the by-product of higher exact boundary match accuracy. 
% We test the model on the Stanford Question Answering Dataset (SQuAD) \cite{rajpurkar2016squad}, which contains a variety of human-generated factoid and non-factoid questions and results in a question distribution closer to that in real-world applications. 
The experiments on the Stanford Question Answering Dataset (SQuAD) \cite{rajpurkar2016squad}, which contains a variety of human-generated factoid and non-factoid questions, have shown the effectiveness of above three contributions.
% on top of a bi-directional GRU \cite{bengio2015deep} passage encoder to 

%kazi edited it
Our paper is organized as follows. We formally define the RCQA problem first. Next, we describe our baseline with a neural network component. We present the end-to-end dynamic chunk reader model next. Finally, we analyze our experimental results and discuss the related work.

\section{Problem Definition}
\label{sec:definition}

\begin{table}
\caption{{Example of questions (with answers) which can be potentially answered with RC on a Wikipedia passage. The first question is factoid, asking for an entity. The second and third are non-factoid.}}
\label{example}
\centering
\scriptsize
\begin{tabular}{|p{8cm}|}
\hline
The United Kingdom (UK) intends to withdraw from the European Union (EU), a process commonly known as Brexit, as a result of a June 2016 referendum in which 51.9\% voted to leave the EU. The separation process is complex, causing political and economic changes for the UK and other countries. As of September 2016, neither the timetable nor the terms for withdrawal have been established: in the meantime, the UK remains a full member of the European Union. The term "Brexit" is a portmanteau of the words "British" and "exit". \\
\hline
Q1. Which country withdrew from EU in 2016?\\% organization did United Kingdom withdraw from? \\
A1. United Kingdom\\ \hline %European Union (EU) \\ \hline
Q2. How did UK decide to leave the European Union? \\
A2. as a result of a June 2016 referendum in which 51.9\% voted to leave the EU \\ \hline
Q3. What has not been finalized for Brexit as of September 2016? \\
A3. neither the timetable nor the terms for withdrawal \\ \hline
\end{tabular}
\end{table}

Table~\ref{example} shows an example of our RC setting where the goal is to answer a question $Q_i$, factoid (Q1) or non-factoid (Q2 and Q3), based on a supporting passage $P_i$, by selecting a continuous sequence of text $A_i \subseteq P_i$ as answer. $Q_i$, $P_i$, and $A_i$ are all word sequences, where each word is drawn from a vocabulary, $V$.
% Without loss of generality, the RCQA problem can be formally defined as the task to understand two given word sequences, a supporting passage, $P_i$, and a question, $Q_i$; and select a continuous sequence of text, $A_i$, from $P_i$ as answer to $Q_i$. Words in the sequences are from a vocabulary, $V$. 
The $i$-th instance in the training set is a triple in the form of $(P_i,Q_i,A_i)$, where $P_i = (p_{i1}, \dots, p_{i|P_i|})$, $Q_i = (q_{i1}, \dots, q_{i|Q_i|})$, and $A_i = (a_{i1}, \dots, a_{i|A_i|})$ ($p_{i\cdot},q_{i\cdot},a_{i\cdot} \in V$). Owing to the disagreement among annotators, there could be more than one correct answer for the same question; and the $k$-th answer to $Q_i$ is denoted by $A^k_i = \{a^k_{i1}, \dots, a^k_{i|A^k_i|}\}$. An answer candidate for the $i$-th training example is defined as $c_i^{m,n}$, a sub-sequence in $P_i$, that spans from position $m$ to $n$ ($1 \leq m \leq n \leq |P_i|$). The ground truth answer $A_i$ could be included in the set of all candidates $C_i= \{c_i^{m,n}\ | \forall m,n \in N^+, subj(m,n,P_i) \; and \; 1 \leq m \leq n \leq |P_i| \}$, where $subj(m,n,P_i)$ is the constraint put on the candidate chunk for $P_i$, such as, ``$c_i^{m,n}$ can have at most 10 tokens'', or ``$c_i^{m,n}$ must have a pre-defined POS pattern''. To evaluate a system's performance, its top answer to a question is matched against the corresponding gold standard answer(s).

\paragraph{Remark: Categories of RC Tasks}
%Different categories of RC tasks were explored besides the above definition.
Other simpler variants of the aforementioned RC task were explored in the past.
For example, \textbf{\emph{quiz-style}} datasets (e.g., MCTest \cite{richardson2013mctest}, MovieQA \cite{tapaswi2015movieqa}) have multiple-choice questions with answer options. \textbf{\emph{Cloze-style}} datesets\cite{hermann2015teaching,hill2015goldilocks,onishi-16}, usually automatically generated, have factoid ``question''s created by replacing the answer in a sentence from the text with blank. For the \textbf{\emph{answer selection}} task this paper focuses on, several datasets exist, e.g. TREC-QA for factoid answer extraction from multiple given passages, bAbI \cite{DBLP:journals/corr/WestonCB14} designed for inference purpose, and the SQuAD dataset \cite{rajpurkar2016squad} used in this paper. To the best of our knowledge, the SQuAD dataset is the only one for both factoid and non-factoid answer extraction with a question distribution more close to real-world applications.

\section{Baseline: Chunk-and-Rank Pipeline with Neural RC}
%\section{Starting with a Baseline Pipeline-Style Chunk-Level Ranking Method}
% \section{A Baseline Ranker Method}
\label{sect_baseline}
%TODO By Mo: can be efficiently trained ? 
In this section we modified a state-of-the-art RC system for cloze-style tasks for our answer extraction purpose, to see how much gap we have for the two type of tasks, and to inspire our end-to-end system in the next section.
In order to make the cloze-style RC system to make chunk-level decision, we use the RC model to generate features for chunks, which are further used in a feature-based ranker like in ~\cite{rajpurkar2016squad}.
% Feature-based ranking systems have shown to be effective for QA tasks \cite{ferrucci2013watson}. To obtain a baseline to compare to our end-to-end neural reading comprehension model, we start off by building a cascaded model as a ranker system. 
As a result, this baseline can be viewed as a deep learning based counterpart of the system in \cite{rajpurkar2016squad}. It has two main components: 1) a stand-alone answer chunker, which is trained to produce overlapping candidate chunks, and 2) a neural RC model, which is used to score each word in a given passage to be used thereafter for generating chunk scores.

% \textit{\textbf{Answer Chunking}} 
\noindent\textbf{Answer Chunking}
% TODO By Mo: ``hard loss'' not mentioned before
% TODO By Mo: no training here?
%A rule-based answer chunker as in Rajpurkar et al.~\shortcite{rajpurkar2016squad} suffers from chunking errors that is not able to be recovered during training. 
To reduce the errors generated by the rule-based chunker in \cite{rajpurkar2016squad}, first, we capture the part-of-speech (POS) pattern of all answer sub-sequences in the training dataset to form a \textit{POS pattern trie tree}, and then apply the answer POS patterns to passage $P_i$ to acquire a collection of all subsequences (chunk candidates) $C_i$ whose POS patterns can be matched to the \textit{POS pattern trie}. This is equivalent to putting an constraint $subj(m,n,P_i)$ to candidate answer chunk generation process that only choose the chunk with a POS pattern seen for answers in the training data. Then the sub-sequences $C_i$ are used as answer candidates for $P_i$. Note that overlapping chunks could be generated for a passage, and we rely on the ranker to choose the best candidate based on features from the cloze-style RC system. Experiments showed that for $>90\%$ of the questions on the development set, the ground truth answer is included in the candidate set constructed in such manner.

\noindent\textbf{Feature Extraction and Ranking} 
For chunk ranking, we (1) use neural RCQA model to annotate each $p_{ij}$ in passage $P_i$ to get score $s_{ij}$, then (2) for every chunk $c_{i}^{m,n}$ in passage $i$, collect scores $(s_{im},\dots, s_{in})$ for all the $(p_{im}, ..., p_{in})$ contained within $c_{i}^{m,n}$, and (3) extract features on the sequence of scores $(s_{im},\dots, s_{in})$ to characterize its scale and distribution information, which serves as the feature representation of $c_{i}^{m,n}$. In step (1) to acquire $s_{ij}$ we train and apply a word-level single-layer Gated Attention Reader \footnote{We tried using more than one layers in Gated Attention Reader, but no improvement was observed.} \cite{dhingra2016gated}, which has state-of-the-art performance on CNN/DailyMail cloze-style RC task. In step (3) for chunk $c_{i}^{m,n}$, we designed 5 features, including 4 statistics on $(s_{im},\dots, s_{in})$: \textit{maximum, minimum, average and sum}; as well as the count of matched POS pattern within the chunk, which serves as an answer prior. We use these 5 features in a state-of-the-art ranker \cite{Ganji:2011:SIGIR}.

\section{Dynamic Chunk Reader}
\label{sect_dcr}
\begin{figure}
\begin{center}
\includegraphics[trim={0.2cm 0 0 0},scale=0.19]{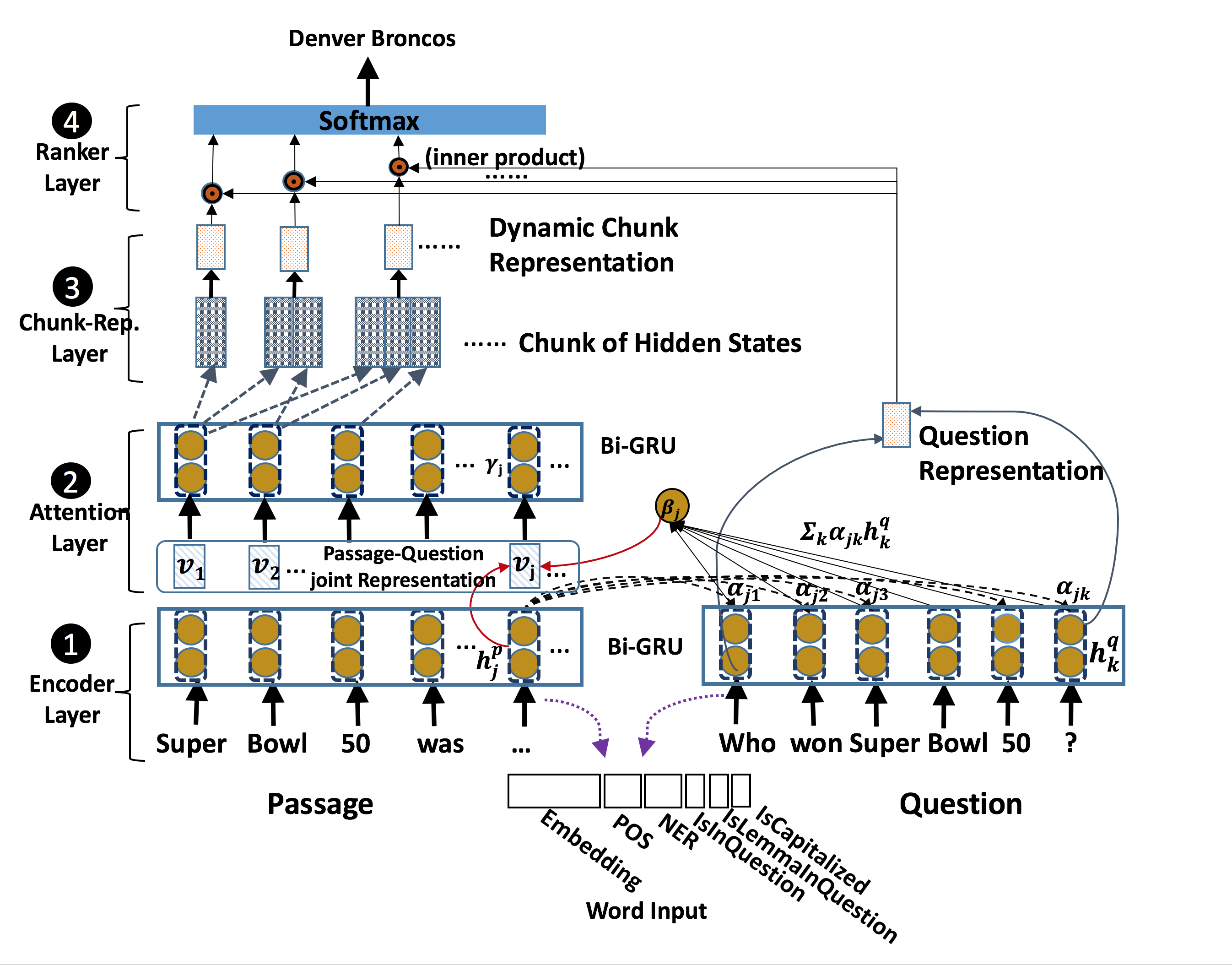}
\caption{{The main components in dynamic chunk reader model (from bottom to top) are bi-GRU encoders for passage and question, a word-by-word attention bi-GRU for passage, dynamic chunk representations that are transformed from pooled dynamic chunks of hidden states, the question attention on every chunk representation and final answer chunk prediction.}}
\label{fig_arch}
\vspace{-0.2in}
\end{center}
\end{figure}

%kazi
The dynamic chunk reader (\dcr) model is presented in Figure~\ref{fig_arch}. Inspired by the baseline we built, DCR is deemed to be superior to the baseline for 3 reasons. First, each chunk has a representation constructed dynamically, instead of having a set of pre-defined feature values. Second, each passage word's representation is enhanced by word-by-word attention that evaluates the relevance of the passage word to the question. Third, these components are all within a single, end-to-end model that can be trained in a joint manner.
%Mo's version - edited by Wei
%The Dynamic Chunk Reader model is presented in Figure \ref{fig_arch}. The key idea is inspired by the baseline system we built. But we believe DCR is superior in four reasons. First the representation of each chunk is dynamically constructed instead of represented as a few predefined feature values. Second, all the candidate chunk representations are learned to take question representation into consideration through a word-by-word attention mechanism. Third, all chunk representations are list-wise compared with question attention. Fourth, all the components are within a single end-to-end model that could be jointly trained, which means the single word representation, the chunk’s representation, the attention and the list-wise comparison among candidate chunks are all automatically adjusted during learning process.

\dcr\ works in four steps.
%By Mo
%kazi 
%\item The {\em encoder layer} uses bidirectional RNNs to encode each passage and question word so that the contextual information is also encoded in a word's representation. 
First, the \textit{\textbf{encoder layer}} encodes passage and question separately, by using bidirectional recurrent neural networks (RNN).
%kazi
%\item The {\em attention layer} uses an additional RNN to make passage word representations more question-related with attention from question based on alignments between representations for passage and question words.
Second, the \textit{\textbf{attention layer}} calculates the relevance of each passage word to the question.
% which is essentially through a two way attention mechanism using an additional RNN for passage words.
% TODO by Mo: for Yang, could you use one sentence to summarize the details of all constrains we have for the chunk extraction here?
%kazi
%\item The {\em chunk extraction layer} dynamically extracts, for each passage word, all possible candidate chunks within  a window size.
%Third, the \textit{\textbf{chunk extraction layer}} dynamically extracts, for each passage word, all possible candidate chunks that start with the word and are within a window size. 
% Each chunk represents a constituent part that might be noted when reading the passage given the question. 
% and let the model differentiate patterns of answers and rank them according to the question.
 %kazi
%\item The {\em chunk representation layer} creates a representation for each chunk by concatenating its first word's state in the forward RNN and its last word's state in the backward RNN.
Third, the \textit{\textbf{chunk representation layer}} dynamically extracts the candidate chunks from the given passage, and create chunk representation that encodes the contextual information of each chunk.
%Fourth, the \textit{\textbf{chunk representation layer}} to construct chunk representations given the internal hidden states of RNNs. We concatenate the last state of forward RNN and first state of backward RNN to represent the question. 
%kazi: commented the footnote out because the point is mentioned later
%\footnote{Other pooling methods can be applied to generate the chunk representation from the representations of all words within the chunk. But  the concatenation method showed best result.}
%kazi
Fourth, the \textit{\textbf{ranker layer}} scores the relevance between the representations of a chunk and the given question, and ranks all candidate chunks using a softmax layer.  
%Fifth, the \textit{\textbf{ranker layer}} to score each chunk, based on the chunk representation and input question representation. This step evaluates the relevance between the question and the context around the chunk, and rank all candidates list-wise using a SoftMax layer.
% Finally we have a softmax node on top of the scores of all the chunks to maximize the probability of correct answer.
% TODO By Mo: Do we have to talk about something like attention as the commented below?
% I think we should talk about the pointer network style attention
% The score means given the question and the context around each chunk that is encoded in the representation of chunk, which chunk is more relevant, important or potential to answer the question. The attention mechanism is similar to the pointer network \cite{vinyals2015pointer} and attention mechanism in AS reader \cite{kadlec2016text}.
We describe each step in details below.

\noindent\textbf{Encoder Layer}
%kazi
%TODO By Mo: two different bi-directional RNN encoders?
We use bi-directional RNN encoder to encode $P_i$ and $Q_i$ of example $i$, and get hidden state for each word position $p_{ij}$ and $q_{ik}$.\footnote{We can have separated parameters for question and passage encoders but a single shared encoder for both works better in the experiments.}
% Each word $p_{ij}$ or $q_{ik}$ in $P_i$ and $Q_i$ of example $i$ are encoded separately by two different bi-directional RNN encoders. 
As RNN input, a word is represented by a row vector $x \in \mathbb{R}^n$. $x$ can be the concatenation of word embedding and word features (see Fig. \ref{fig_arch}). The word vector for the $t$-th word is $x_t$. A word sequence is processed using an RNN encoder with gated recurrent units (GRU) \cite{bengio2015deep}, which was proved to be effective in RC and neural machine translation tasks \cite{bahdanau2014neural,kadlec2016text,dhingra2016gated}. For each position $t$, GRU computes $h_t$ with input $x_t$ and previous state $h_{t-1}$, as:
%The input to the encoding phase is a sequence of words, drawn from a vocabulary $V$. Each  $p_{ij}$ or $q_{ik}$ is represented by a continuous-valued word vector $x \in \mathbb{R}^n$ stored into a 2 dimensional matrix $X \in \mathbb{R}^{|V| \times n}$. $x$ could be a concatenation of feature vectors for each word. The vector sequence $X$ is fed into a Recurrent Neural Networks encoder with Gated Recurrent Units (GRU)  \cite{bengio2015deep}, which was proved to be an effective recurrent neural networks in neural reading comprehension or neural machine translation models \cite{bahdanau2014neural,kadlec2016text,dhingra2016gated}. For each time step $t$ in the input sequence, the GRU takes as input the word embedding $x_t$ and evolves a hidden state $h_{t-1}$ to $h_t=f(x_t, h_{t-1})$, where $f$ is defined as:
\begin{eqnarray}
  r_t &=&\sigma(W_rx_t+U_rh_{t-1}) \\
  u_t &=&\sigma(W_ux_t+U_uh_{t-1}) \\
  \bar{h_t}&=&tanh(Wx_t+U(r_t\odot h_{t-1})) \\
  h_t&=&(1-u_t)\cdot h_{t-1} + u_t \cdot \bar{h_t}
\end{eqnarray}
%kazi
where $h_t$, $r_t$, and $u_t \in \mathbb{R}^d$ are d-dimensional hidden state, reset gate, and update gate, respectively; $W_{\{r,u\}}$, $W \in \mathbb{R}^{n\times d}$ and $U_{\{r,u\}}$, $U \in \mathbb{R}^{d\times d}$ are the parameters of the GRU; $\sigma$ is the sigmoid function, and $\odot$ denotes element-wise production. For a word at $t$, we use the hidden state $\overrightarrow{h}_t$ from the forward RNN as a representation of the preceding context, and the $\overleftarrow{h}_t$ from a backward RNN that encodes text reversely, to incorporate the context after $t$. Next, $h_t=[\overrightarrow{h_t};\overleftarrow{h_t}]$, the bi-directional contextual encoding of $x_t$, is formed. $[\cdot ;\cdot]$ is the concatenation operator. To distinguish hidden states from different sources, we denote the $h_j$ of $j$-th word in $P$ and the $h_k$ of $k$-th word in $Q$ as $h^p_{j}$ and $h^q_{k}$ respectively.
%Kazi: removed the phrase below
%which parameters are shared.
%Where $h_t$, $r_t$ and $u_t \in \mathbb{R}^h$ are the recurrent state, the reset gate and update gate respectively. $W_{\{r,u\}}$, $W \in \mathbb{R}^{h\times d}$, $U_{\{r,u\}}$, $U \in \mathbb{R}^{h\times h}$ are the parameters of the GRU, $\sigma$ is the sigmoid function and $\odot$ is the element-wise production. We use the hidden state $h_t$ as a representation of the word $x_t$ in the context of the preceding sequences input $x_{<t}$. To incorporate context of future tokens for word $x_t$, we use a backward RNN by input the sequence in reverse to a normal RNN. Then we concatenate the hidden state of $\overleftarrow{h_t}$ of backward RNN with the hidden state of $\overrightarrow{h_t}$ of forward RNN as the final representation of the word $x_t$ in the complete context of the passage, which is denoted as $h_t=[\overrightarrow{h_t},\overleftarrow{h_t}]$. We will denote by $h^p_{i}$ and $h^q_{j}$ as the contextual encodings for word $i$ in document $P$ and word $j$ in query $Q$ and respectively.

\noindent\textbf{Attention Layer}
%kazi
%Attention mechanism \cite{nmt,image_captioning} allows a model to store query-specific information in the representation of each word in an input sequence. Motivated by this, we propose a novel attention mechanism inspired by word-by-word style attention methods \cite{snli,santos2016attentive}. The key idea is to compute similarity scores between question-passage hidden state pairs, and derive question-related passage representations weighted by these scores.
Attention mechanism in previous RC tasks \cite{kadlec2016text,hermann2015teaching,sordoni2016iterative,dhingra2016gated,cui2016aoa,cui2016consensus} enables question-aware passage representations. We propose a novel attention mechanism inspired by word-by-word style attention methods \cite{rocktaschel2015reasoning,wang2015learning,santos2016attentive}. For each $p_j$, a question-attended representation $v_j$ is computed as follows (example index $i$ is omitted for simplicity):
% 
% Attentive pooling (AP) \cite{santos2016attentive} is  an  approach  that  enables  the  pooling  layer  to  be  aware  of  the  current  input  pair,  in  a  way that information from the question q can directly influence the computation of the passage representation, and vice versa. The main idea consists of learning a similarity measure over the input pairs, and uses the similarity scores between the segments to compute attention vectors. When AP is applied to biRNN, the network learns the similarity measure over the hidden states produced by the biRNN when processing the two input sequences. Inspiring from this idea, we proposed an approach to learn the similarity measure and this measure affects input pair's representation in both ways.
\begin{eqnarray}
\alpha_{jk} &=& h^p_j \cdot h^q_k, \ \\
\beta_j &=&\sum_{k=1}^{|Q|}{\alpha_{jk}h^q_k} \\
%p_j^{\alpha}=\sum_{i}{\alpha_{ij}h^p_i} \\
%x^{q}_j=[h^q_j : p_j^{\alpha}] \\
v_j &=& [h^p_j ; \beta_j]
\end{eqnarray}
%kazi
%Specifically, given all the hidden states produced by the bottom layer biRNNs for passage and question words (see Figure~\ref{fig_arch}), we compute the similarity between every pair of $i$-th passage and $j$-th question word as the inner product between $h^p_i$ and $h_{q_j}$. This yields a similarity matrix, $\alpha$, of size $|P| * |Q|$. Next, for the $i$-th passage word, we do a weighted average pooling over the question dimension to obtain $q_i^\alpha$. The weighted average pooling over question words for the $i$-th passage word results in a question representation for the word. Finally, we concatenate this question representation together with the passage word's hidden states produced by the biRNN below to form the new representation for the passage word, $x'^{p}_i$.\footnote{We also tried to follow the idea of two-way attention and have similar passage representation input to question side as well. However, this does not lead to improved accuracy due to the confusion created by long passages in RC. Consequently, we have this representation layer only on the passage side.} An additional biRNN then takes $x'^{p}_i$s as inputs and produces a question-related hidden state, $h^{\alpha}_{i}$.
where $h_j^p$ and $h_k^q$ are hidden states from the bi-directional RNN encoders (see Figure~\ref{fig_arch}). An inner product, $\alpha_{jk}$, is calculated between $h_j^p$ and every question word $h_k^q$. It indicates how well the passage word $p_j$ matches with every question word $q_k$. $\beta_j$ is a weighted pooling of $|Q|$ question hidden states, which serves as a $p_j$-aware question representation. The concatenation of $h_j^p$ and $\beta_j$ leads to a passage-question joint representation, $v_j \in \mathbb{R}^{4d}$.\footnote{We tried another word-by-word attention methods as in \cite{santos2016attentive}, which has similar passage representation input to question side. However, this does not lead to improvement due to the confusion caused by long passages in RC.
Consequently, we used the proposed simplified version of word-by-word attention on passage side only.} 
% Consequently, we have this representation layer only on the passage side.}
Next, we apply a second bi-GRU layer taking the $v_j$s as inputs,
% Note that $\gamma_j$ is also a bi-directional representation where
and obtain forward and backward representations $\overrightarrow{\gamma_j}$ and $\overleftarrow{\gamma_j}$ $\in \mathbb{R}^d$, and in turn their concatenation, $\gamma_j=[\overrightarrow{\gamma_j};\overleftarrow{\gamma_j}]$.

\noindent\textbf{Chunk Representation Layer}
%TODO By Mo: how it built with consideration of input question?
% kazi edited it
A candidate answer chunk representation is dynamically created given attention layer output. We first decide the text boundary for the candidate chunk, and then form a chunk representation using all or part of those $\gamma_j$ outputs inside the chunk. To decide a candidate chunk (boundary): we tried two ways: (1) adopt the \textit{POS trie}-based approach used in our baseline, and (2) enumerate all possible chunks up to a maximum number of tokens. For (2), we create up to $N$ (max chunk length) chunks starting from any position $j$ in $P_j$. Approach (1) can generate candidates with arbitrary lengths, but fails to recall candidates whose POS pattern is unseen in training set; whereas approach (2) considers all possible candidates within a window and is more flexible, but over-generates invalid candidates.

%Kazi: commented the part below and rephrased it above
%Surprisingly, both approaches achieve similar performance on chunk ranker. Using \textit{POS Trie} is more efficient since the number of candidates to consider is less than number of all enumerated candidates (for a working N $>=$10), and there's no limitation on chunk length; whereas approach (2) is able to include chunks that \textit{POS Trie} can not catch, no POS is needed, but could include invalid chunks .

%TODO By Mo: I do not understand this sentence "as the representation of the chunk of words corresponding to the tag sequence starting at current word position" so I did not use it in this version.
% kazi edited it
For a candidate answer chunk $c^{m,n}$ spanning from position $m$ to $n$ inclusively, we construct chunk representation $\overline \gamma_{m,n} \in \mathbb{R}^{2d}$ using every $\gamma_j$ within range $[m,n]$, with a function $g(\cdot)$. Formally,
%By Mo: I moved the equation in this subsection
\begin{eqnarray}
% C=g(h^{\alpha}_{i1}, h^{\alpha}_{i2}, ..., h^{\alpha}_{i|C|}) \nonumber
\overline \gamma_{m,n}=g(\gamma_m, \dots, \gamma_n) \nonumber
%kazi: removed the _i from C since we denote the chunk by C
\end{eqnarray}
We experimented with several pooling functions (e.g., max, average) for $g(\cdot)$, and found out that, instead of pooling, the best function is to concatenate the hidden state of the first word in a chunk in forward RNN and that of the last word in backward RNN. Formally,
\begin{eqnarray}
\overline \gamma_{m,n}=g(\gamma_{m}, \dots, \gamma_{n}) = [\overrightarrow{\gamma_{m}} ; \overleftarrow{\gamma_{n}}]
% C=g(h^{\alpha}_{i1}, h^{\alpha}_{i2}, ..., h^{\alpha}_{i|C|}) = [\overrightarrow{h^{\alpha}_{i1}} : \overleftarrow{h^{\alpha}_{i|C|}}]
%kazi: removed the _i from C since we denote the chunk by C
\end{eqnarray}
% kazi edited it
We hypothesize that the hidden states at that two ends can better represent the chunk's contexts, which is critical for this task, than the states within the chunk. This observation also agrees with \cite{kobayashi2016dynamic}.

\noindent\textbf{Ranker Layer}
%TODO By Mo: I am not quite sure about the maths in eq(9), I will check the code after I finish this pass
Each chunk $c^{m,n}$ is evaluated on its context similarity to the question, by taking the cosine similarity between the chunk context representation $\bar\gamma_{m,n}$ acquired from chunk representation layer, and the question representation which is the concatenation of the last hidden state in forward RNN and the first hidden state in backward RNN. Thus, for training example $i$, we have the probability of the chunk $c^{m,n}_i$ as
\begin{eqnarray}
\mathbb{P}(c^{m,n}_i|P_i, Q_i) = softmax(\overline \gamma^i_{m,n} \cdot [\overrightarrow{h^{Q_i}_{|Q_i|}} ; \overleftarrow{h^{Q_i}_1}])
\end{eqnarray}
where $\bar\gamma^i_{m,n}$ denotes representation of the chunk $c^{m,n}_i$, $\overrightarrow{h^{Q_i}_k}$ or $\overleftarrow{h^{Q_i}_k}$ is the $k$-th hidden state output from question $Q_i$'s forward and backward RNN encoder, respectively. % The attention is similar to attention mechanism used in attention sum reader \cite{kadlec2016text}, which is a inner product between dynamic chunk representation and question representation. The question representation is a concatenation of hidden state of last word in forward RNN and hidden state of first word in backward RNN. 
In runtime, the chunk with the highest probability is taken as the answer. In training, the following negative log likelihood is minimized:
\begin{eqnarray}
\mathbb{L} = -\sum_{i=1}^N \log\mathbb{P}(A_i|P_i, Q_i)
% \mathbb{L} = -\sum_{i=1}^NI(A_i\in C_i)*log\mathbb{P}(A_i|P_i, Q_i)
\end{eqnarray}
Note that the $i$-th training instance is only used when $A_i$ is included in the corresponding candidate chunk set $C_i$, i.e. $\exists_{m,n} A_i = c^{m,n}_i$.
% where $I(\cdot)$ is the indicator function returning 1 if the ground truth $A_i$ is included in the candidate chunk set $C_i$, or 0 otherwise. 
The softmax in the final layer serves as the list-wise ranking module similar in spirit to \cite{cao2007learning}.
% We calculate the attention from the question to every chunk. The question representation is the concatenation of hidden state of last word in forward RNN and that of first word in backward RNN. The attention is similar to attention mechanism used in attention sum reader \cite{kadlec2016text}, which is a inner product between dynamic chunk representation and question representation. The question representation is a concatenation of hidden state of last word in forward RNN and hidden state of first word in backward RNN. Given the attention score, we use softmax to directly pick the best candidate as the answer chunk prediction.

\section{Experiments}

\paragraph{Dataset}
We used the Stanford Question Answering Dataset (SQuAD) \cite{rajpurkar2016squad} for the experiment. SQuAD came into our sight because it is a mix of factoid and non-factoid questions, a real-world data (crowd-sourced), and of large scale (over 100K question-answer pairs collected from 536 Wikipedia articles). Answers range from single words to long, variable-length phrase/clauses. It is a relaxation of assumptions by the cloze-style and quiz-style RC datasets in the Problem Definition section.
% as a system must consider, for this dataset, up to $O(n^2)$ text chunks from a given passage with $n$ words. 
% We used the Stanford Question Answering Dataset (SQuAD) \cite{rajpurkar2016squad} is a crowdsourced dataset, which is unique in the sense that it has a mix of factoid and non-factoid questions. With over 100K question-answer pairs collected from 536 Wikipedia articles, this dataset has questions with answers ranging from a named entity to long, variable-length phrase/clauses. Consequently, it overcomes some of the shortcomings of the aforementioned datasts as a system must consider, for this dataset, up to $O(n^2)$ text chunks from a given passage with $n$ words. 

\noindent\textbf{{Features}}
% kazi edited it
The input vector representation of each word $w$ to encoder RNNs has six parts including a pre-trained 300-dimensional GloVe embedding \cite{pennington2014glove} and five features (see Figure~\ref{fig_arch}): (1) a one-hot encoding (46 dimensions) for the part-of-speech (POS) tag of $w$; (2) a one-hot encoding (14 dimensions) for named entity (NE) tag of $w$; (3) a binary value indicating whether $w$'s surface form is the same to any word in the quesiton; (4) if the lemma form of $w$ is the same to any word in the question; and (5) if $w$ is caplitalized. Feature (3) and (4) are designed to help the model align the passage text with question. Note that some types of questions (e.g., ``who'', ``when'' questions) have answers that have a specific POS/NE tag pattern. For instance, ``who'' questions mostly have proper nouns/persons as answers and ``when'' questions may frequently have numbers/dates (e.g., a year) as answers. Thus, we believe that the model could exploit the co-relation between question types and answer POS/NE patterns easier with POS and NE tag features. %For every type of questions, like when, how, why etc., we would expect that the answer has POS tag sequence related to certain patterns. For example, if the question is about how many/much, the answer POS highly like is a number (CD); while if the question is about why, the pattern might be starting with a verb phrase, as the answer needs to explain something. By using the POS features we hope the model automatically learn the relations between the question types and POS patterns. 

\noindent\textbf{Implementation Details}
%kazi edited it
We pre-processed the SQuAD dataset using Stanford CoreNLP tool\footnote{ stanfordnlp.github.io/CoreNLP/} \cite{manning-EtAl:2014:P14-5} with its default setting to tokenize the text and obtain the POS and NE annotations. To train our model, we used stochastic gradient descent with the ADAM optimizer \cite{kingma2014adam}, with an initial learning rate of 0.001. All GRU weights were initialized from a uniform distribution between (-0.01, 0.01). The hidden state size, $d$, was set to 300 for all GRUs. The question bi-GRU shared parameters with the passage bi-GRU, while the attention-based passage bi-GRU had its own parameters. We shuffled all training examples at the beginning of each epoch and adopted a curriculum learning approach \cite{bengio2009curriculum}, by sorting training instances by length in every 10 batches, to enable the model start learning from relatively easier instances and to harder ones.
% Curriculum learning has been proved to be both time efficient in training and help the model to learn from relatively easier instances from harder instances which is very similar to how human learns. 
We also applied dropout of rate 0.2 to the embedding layer of input bi-GRU encoder, and gradient clipping when the norm of gradients exceeded 10. We trained in mini-batch style (mini-batch size is 180) and applied zero-padding to the passage and question inputs in each batch. We also set the maximum passage length to be 300 tokens, and pruned all the tokens after the 300-th token in the training set to save memory and speed up the training process. This step reduced the training set size by about 1.6\%. During test, we test on the full length of passage, so that we don't prune out the potential candidates. We trained the model for at most 30 epochs, and in case the accuracy did not improve for 10 epochs, we stopped training. 

For the feature ranking-based system, we used jforest ranker \cite{Ganji:2011:SIGIR} with LambdaMART-RegressionTree algorithm and the ranking metric was NDCG@10. For the Gated Attention Reader in baseline system, we replicated the method and use the same configurations as in \cite{dhingra2016gated}.

\begin{table}
%\scriptsize
\caption{Results on the SQuAD dataset.}
\label{tbl_results}
\centering
\begin{tabular}{lcccc}
\hline
&\multicolumn{2}{c}{Dev}&\multicolumn{2}{c}{Test}\\
Models&EM&F1&EM&F1\\
\hline
Rajpurkar 2016&39.8\%&51.0\%&40.4\%&51.0\%\\
Wang 2016&59.1\%&70.0\%&59.5\%&70.3\%\\
%DCR &62.0\%&71.2\%&60.3\%&69.8\%\\
%the line below is post submission results
DCR &62.5\%&71.2\%&62.5\%&71.0\%\\

\hline
\end{tabular}
\end{table}

% Do we need a table to show how we achieve the SOTA : how each component helps
\begin{table}
%\scriptsize
\caption{Detailed system experiments on the SQuAD development set.}
\label{tbl_ablation}
\centering
\begin{tabular}{lcc}
\hline
Models&EM&F1\\
\hline
Chunk-and-Rank Pipeline Baseline &49.7\%&64.9\%\\
% Ranker (2 layer)&49.7\%&64.9\%\\
\hline
DCR&62.0\%&71.2\%\\
%Dynamic Chunk Reader (DCR) w/o Chunk-level Ranker & 49.25\%& 56.05\%\\
DCR w/o Word-by-Word Attention &57.6\%&68.7\%\\
DCR w/o POS feature (1) &59.2\%&68.8\%\\
DCR w/o NE feature (2) &60.4\%&70.2\%\\
DCR w/o Question-word feature (3) &59.5\% &69.0\%\\
DCR w/o Question-lemma feature (4) & 61.2\%&69.9\%\\
DCR w/o Capitalized feature (5) & 61.5\%&70.6\%\\
\hline
DCR w POS-trie &62.1\% &70.8\% \\
\hline
\end{tabular}
\end{table}

\noindent\textbf{Results}
%TODO By Mo: I think we could look at all results in Table2 to see whether we should add more discussions to the ablation tests.
Table~\ref{tbl_results} shows our main results on the SQuAD dataset. Compared to the scores reported in \cite{wang2016machine}, our exact match (EM) and F1 on the development set and EM score on the test set are better, and F1 on the test set is comparable. We also studied how each component in our model contributes to the overall performance. Table~\ref{tbl_ablation} shows the details as well as the results of the baseline ranker.
As the first row of Table~\ref{tbl_ablation} shows, our baseline system improves 10\% (EM) over Rajpurkar et al.~\shortcite{rajpurkar2016squad} (Table~\ref{tbl_results}, row 1), the feature-based ranking system. 
However when compared to our DCR model (Table~\ref{tbl_ablation}, row 2), the baseline (row 1) is more than 12\% (EM) behind even though it is based on the state-of-the-art model for cloze-style RC tasks. This can be attributed to the advanced model structure and end-to-end manner of DCR.
% Such large performance gap also demonstrates the big difference between the cloze-style RC tasks and the general RC task that this paper studies; as the state-of-the-art method in the former setting does not work for the latter.
% demonstrating the effectiveness of the features learned from neural RC model, and the proposed POS-trie template.
% This difference can be attributed to both the improved chunking approach with high recall POS-trie, which allows overlapping chunks, and the chunk scoring approach that uses the state-of-the-art neural networks model for cloze-style RC tasks.
% When compared to our DCR model (Table~\ref{tbl_ablation}, row 2), the baseline (row 1) is more than 12\% (EM) behind. Such large performance gap demonstrates the big difference between the cloze-style RC tasks and the general RC task that this paper studies; and gives an evidence that our task is more challenging.
% 's results where we use Gated Attention with one layer. As explained in Section \ref{sect_baseline}, this model incorporates the contributions from state-of-the-art neural RC systems for the cloze-style tasks and improves \~10\%  over the best non-deep-learning based feature ranking system. 

We also did ablation tests on our DCR model. First, replacing the word-by-word attention with Attentive Reader style attention \cite{hermann2015teaching} decreases the EM score by about 4.5\%, showing the strength of our proposed attention mechanism. 
% This results are still significantly better than the baseline results, showing the advantage of our attention method and features, as well as of end-to-end training
Second, we remove the features in input to see the contribution of each feature. The result shows that POS feature (1) and question-word feature (3) are the two most important features.
%By Mo: it is fine for me if we do not have the DCR w/o all features
% Third, we replace the chunk-level ranker layer with a sequential labeling layer results in a huge performance drop.
Finally, combining the DCR model with the proposed POS-trie constraints yields a score similar to the one obtained using the DCR model with all possible $n$-gram chunks. The result shows that (1) our chunk representations are powerful enough to differentiate even a huge amount of chunks when no constraints are applied; and (2) the proposed POS-trie reduces the search space at the cost of a small drop in performance.

\noindent\textbf{Analysis}
To better understand our system, we calculated the accuracy of the attention mechanism of the gated attention reader used in our deep learning-based baseline. We found that it is 72\% accurate i.e., 72\% of the times a word with the highest attention score is inside the correct answer span. This means that, if we could accurately detect the boundary around the word with the highest attention score to form the answer span, we could achieve an accuracy close to 72\%.
In addition, we checked the answer recall of our candidate chunking approach. When we use a window size of 10, 92\% of the time, the ground truth answer will be included in the extracted Candidate chunk set. 
Thus the upper bound of the exact match score of our baseline system is around 66\% (92\% (the answer recall) $\times$ 72\%). From the results, we see our DCR system's exact match score is at 62\%. This shows that DCR is proficient at differentiating answer spans dynamically.

%TODO By Mo: is it possible to merge Fig 2(a)&(b), by turning absolute numbers in (a) to ratios
\begin{figure}
\begin{center}
\subfloat[]{
\includegraphics[scale=0.255]{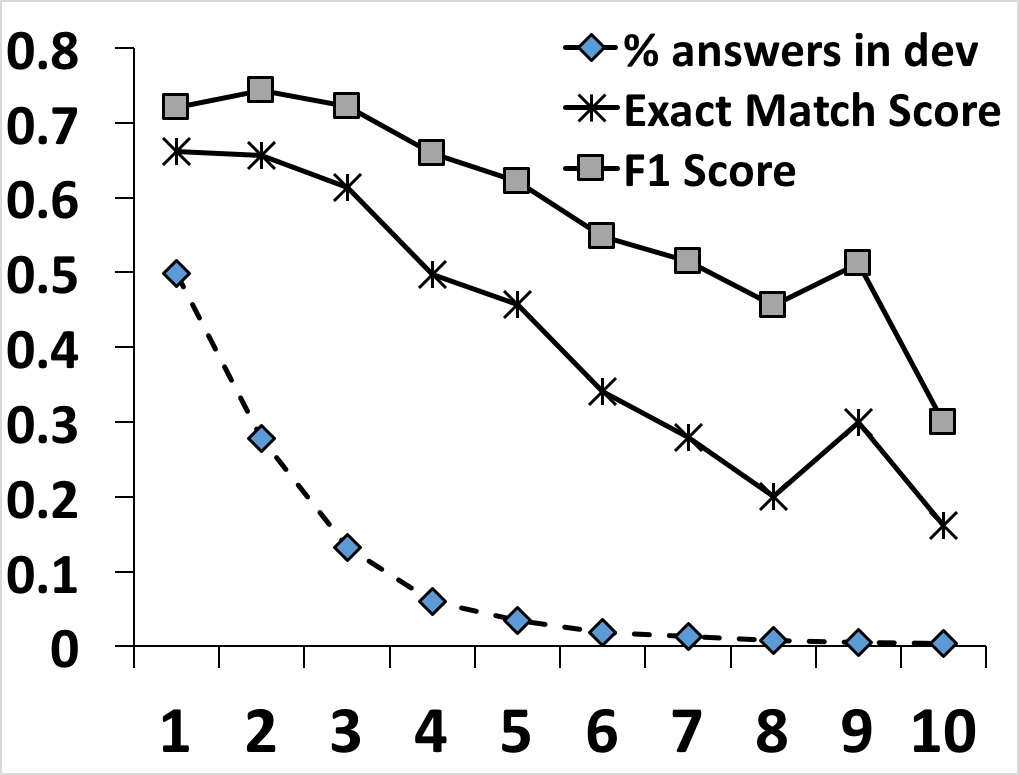}
}
\subfloat[]{
\includegraphics[scale=0.255]{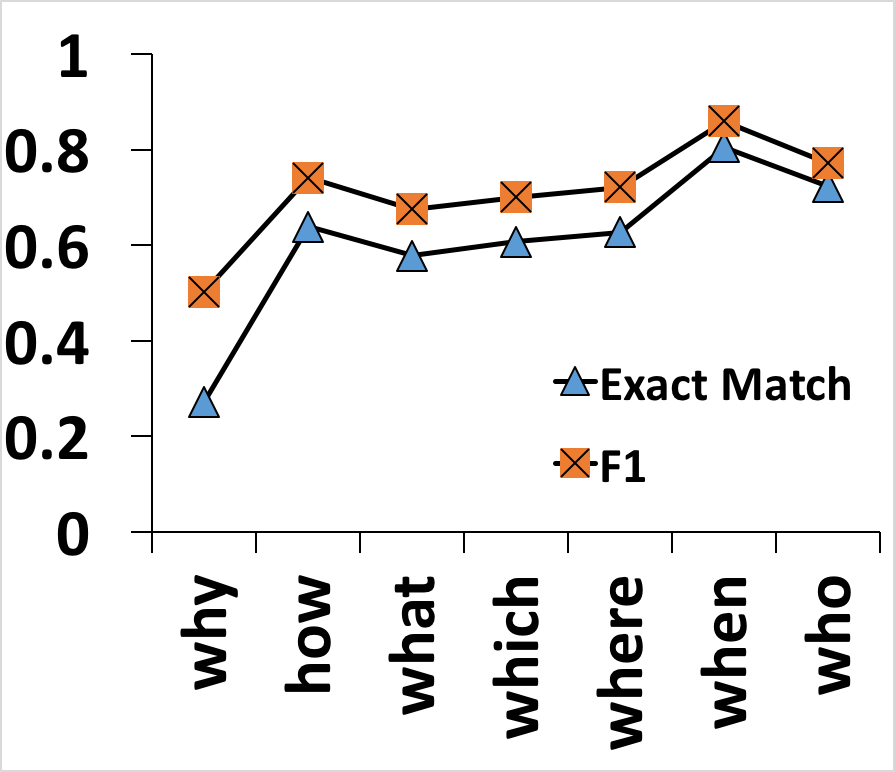}
}
\caption{(a) Variations of DCR performance on ground truth answer length (up to 10) in the development set. The curve with diamond knots also shows the percentage of answers for each length in the development set. (b) Performance comparisons for different question head word.}
\label{fig_stat}
\end{center}
\end{figure}

To further analyze the system's performance while predicting answers of different lengths, we show the exact match (EM) and F1 scores for answers with lengths up to 10 tokens in Figure~2(a). From the graph, we can see that, with the increase of answer length, both EM and F1 drops, but in different speed. The gap between F1 and exact match also widens as answer length increases. However, the model still yields a decent accuracy when the answer is longer than a single word. Additionally, Figure~2(b) shows that the system is better at ``when'' and ``who'' questions, but performs poorly on ``why'' questions. The large gap between exact match and F1 on ``why'' questions means that perfectly identifying the span is harder than locating the core of the answer span.

\begin{figure}
\begin{center}
\includegraphics[scale=0.37]{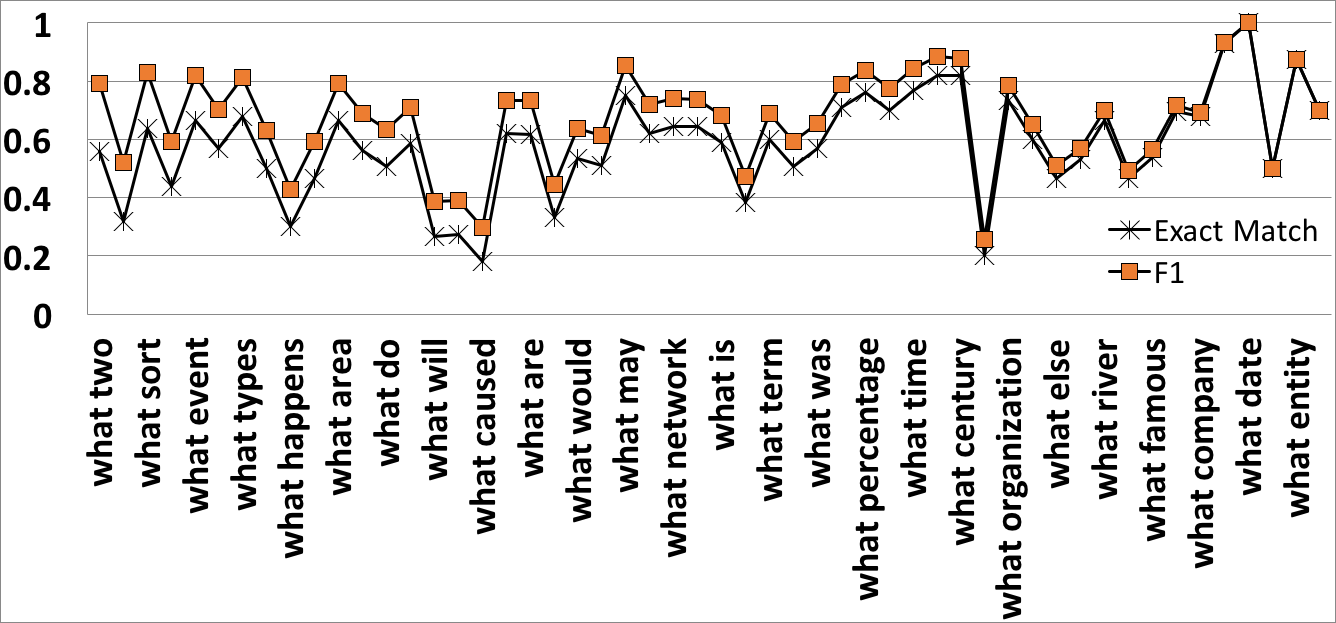}
\caption{Development set performance comparisons for different types of ``what'' questions (considering the types with more than 20 examples in the development set).}
% \label{fig:whatdetail}
\label{whatdetail}
\end{center}
\end{figure}

Since ``what'', ``which'', and ``how'' questions contain a broad range of question types, we split them further based on the bigram a question starts with, and Figure~\ref{whatdetail} shows the breakdown for ``what'' questions. We can see that ``what'' questions asking for explanations such as ``what happens'' and ``what happened'' have lower EM and F1 scores. In contrast, ``what'' questions asking for year and numbers have much higher scores and, for these questions, exact match scores are close to F1 scores, which means chunking for these questions are easier for DCR.

\section{Related Work}
\label{sect_related}

Attentive Reader was the first neural model for factoid RCQA \cite{hermann2015teaching}. It uses Bidirectional RNN (Cho et al., 2014; Chung et al.,2014) to encode document and query respectively, and use query representation to match with every token from the document. Attention Sum Reader \cite{kadlec2016text} simplifies the model to just predicting positions of correct answer in the document and the training speed and test accuracy are both greatly improved on the CNN/Daily Mail dataset. \cite{chen2016thorough} also simplified Attentive Reader and reported higher accuracy. Window-based Memory Networks (MemN2N) is introduced along with the CBT dataset \cite{hill2015goldilocks}, which does not use RNN encoders, but embeds contexts as memory and matches questions with embedded contexts. 
Those models' mechanism is to learn the match between answer context with question/query representation. In contrast, memory enhanced neural networks  like Neural Turing Machines \cite{graves2014neural} and its variants \cite{zhang2015structured,gulcehre2016dynamic,zaremba2015reinforcement} were also potential candidates for the task, and Gulcehre et al. \shortcite{gulcehre2016dynamic} reported results on the bAbI task, which is worse than memory networks. Similarly, sequence-to-sequence models were also used \cite{yu2015empirical,hermann2015teaching}, but they did not yield better results either. 

%\noindent \textbf{Neural RC with Word-by-Word Attention}
Recently, several models have been proposed to enable more complex inference for RC task. For instance, gated attention model \cite{dhingra2016gated} employs a multi-layer architecture, where each layer encodes the same document, but the attention is updated from layer to layer. EpiReader \cite{trischler2016natural} adopted a joint training model for answer extractor and reasoner, where the extractor proposes top candidates, and the reasoner weighs each candidate by examining entailment relationship between question-answer representation and the document. An iterative alternating attention mechanism and gating strategies were proposed in \cite{sordoni2016iterative} to optimize the attention through several hops. 
In contrast, Cui et al.~\shortcite{cui2016aoa,cui2016consensus} introduced fine-grained document attention from each question word and then aggregated those attentions from each question token by summation with or without weights. This system achieved the  state-of-the-art score on the CNN dataset.
Those different variations all result in roughly 3-5\% improvement over attention sum reader, but none of those could achieve higher than that.
Other methods include using dynamic entity representation with max-pooling \cite{kobayashi2016dynamic} that aims to change entity representation with context, and Weissenborn's~\shortcite{weissenborn2016separating} system, which tries to separate entity from the context and then matches the question to context, scoring an accuracy around 70\% on the CNN dataset. 

However, all of those models assume that the answers are single tokens. This limits the type of questions the models can answer. Wang and Jiang~\shortcite{wang2016machine} proposed a match-lstm and achieved good results on SQuAD. However, this approach predicts a chunk boundary or whether a word is part of a chunk or not. In contrast, our approach explicitly constructs the chunk representations and similar chunks are compared directly to determine correct answer boundaries.

\section{Conclusion}
In this paper we proposed a novel neural reading comprehension model for question answering. Different from the previously proposed models for factoid RCQA, the proposed model, dynamic chunk reader, is not restricted to predicting a single named entity as an answer or selecting an answer from a small, pre-defined candidate list. Instead, it is capable of answering both factoid and non-factoid questions as it learns to select answer chunks that are suitable for an input question. DCR achieves this goal with a joint deep learning model enhanced with a novel attention mechanism and five simple yet effective features. Error analysis shows that the DCR model achieves good performance, but still needs to improve on predicting longer answers, which are usually non-factoid in nature.

% The potential future directions could be: 1) a new way of matching chunk type with the pre-defined question type, for the purpose of candidate filtering; and 2) leverage world knowledge for inferring common sense knowledge. We think using open domain knowledge base like DBPedia or Freebase, will be helpful; and 3) enhance neural RC model with better reasoning capability which could be enabled by better representation learned from improved network structure.

\begin{quote}
\begin{small}
\bibliographystyle{aaai}
\bibliography{refer}
\end{small}
\end{quote}

\end{document}